%% file: root.tex
\documentclass[letterpaper, 10 pt, journal, twoside]{IEEEtran} 
\usepackage{amsmath}
\usepackage{listings}
\usepackage{verbatim}
\usepackage{textcomp}
\usepackage{balance}
\usepackage{pgfplots}
\usepackage{textcomp}

\newcommand{\figref}[1]{Fig. \ref{#1}}
\newcommand{\secref}[1]{Sec. \ref{#1}}

\colorlet{RedLine}{red!90}
\colorlet{RedFill}{red!30}

\colorlet{BlueLine}{blue!90}
\colorlet{BlueFill}{blue!35}

\colorlet{BrownLine}{brown!50!black}
\colorlet{BrownFill}{brown!35}

\colorlet{LimeLine}{lime!50!black}
\colorlet{LimeFill}{lime!35}

\input{config}
\begin{document}

\lstset { %
    language=C++,
    backgroundcolor=\color{black!5}, % set backgroundcolor
    basicstyle=\scriptsize,% basic font setting
}

\title{Impact of ROS 2 Node Composition in Robotic Systems}

%%%%%%%%%%%%%%%%%%%%%%%%%%%%%%%%%%%%%%%%%%%%%%%%%%%%%%%%%%%%%%%%

% What are the publishable contribution(s)?
    % description of ros2 composition in literature
    % Benchmarking on a deployed real system, not just toy examples (but also those)
    % Benchmarking performance characteristics of systems with, without, and different types of composition on different system types, in most current code for LTS humble

% https://discourse.ros.org/t/nav2-composition/22175/10

\author{Steve Macenski$^{1}$, Alberto Soragna$^{2}$, Michael Carroll$^{3}$, and Zhenpeng Ge$^{4}$%
%\thanks{Manuscript received: Feb, 2, 2023; Revised April, 20, 2023; Accepted May, 16, 2023.}%Use only for final RAL version
%\thanks{This paper was recommended for publication by Editor  Tamim Asfour upon evaluation of the Associate Editor and Reviewers' comments.
%This work was supported by Samsung Research, iRobot, and Open Robotics.} %Use only for final RAL version
\thanks{$^{1}$Steve Macenski is with Samsung Research America, Mountain View, California
        {\tt\footnotesize stevenmacenski@gmail.com}}%
\thanks{$^{2} $Alberto Soragna is with iRobot, Bedford, Massachusetts
        {\tt\footnotesize asoragna@irobot.com}}%
\thanks{$^{3} $Michael Carroll is with Open Robotics, Mountain View, California
        {\tt\footnotesize michael@openrobotics.org}}%
\thanks{$^{3} $Zhenpeng Ge is with University of Electronic Science and Technology of China, Sichuan, China
        {\tt\footnotesize gezhenpeng@std.uestc.edu.cn}}%
%\thanks{Digital Object Identifier (DOI): see top of this page.}
}

% \author{
% \authorblockN{Steve Macenski}
% \authorblockA{
% \textit{Samsung Research America}\\
% s.macenski@samsung.com}
% \and
% \authorblockN{Alberto Soragna}
% \authorblockA{
% \textit{iRobot}\\
% asoragna@irobot.com}
% \and
% \authorblockN{Michael Carroll}
% \authorblockA{
% \textit{Open Robotics}\\
% michael@openrobotics.org}
% \and 
% \authorblockN{Zhenpeng Ge}
% \authorblockA{
% \textit{UESTC}\\
% gezhenpeng@std.uestc.edu.cn}
% }

% who, what, when, where, why, how
% be linear in my thought train and explain fully 

\maketitle
\begin{abstract}

The Robot Operating System 2 (ROS 2) is the second generation of ROS representing a step forward in the robotic framework.
Several new types of nodes and executor models are integral to control where, how, and when information is processed in the computational graph.
This paper explores and benchmarks one of these new node types - the \textit{Component} node - which allows nodes to be composed manually or dynamically into processes while retaining separation of concerns in a codebase for distributed development. 
Composition is shown to achieve a high degree of performance optimization, particularly valuable for resource-constrained systems and sensor processing pipelines, enabling distributed tasks that would not be otherwise possible in ROS~2.

In this work, we briefly introduce the significance and design of node composition, then our contribution of benchmarking is provided to analyze its impact on robotic systems.
Its compelling influence on performance is shown through several experiments on the latest Long Term Support (LTS) ROS~2 distribution, Humble Hawksbill.

\end{abstract}


\section{Introduction}
\label{sec:introduction}

% general topic intro
\IEEEPARstart{T}{he} Robot Operating System (ROS) is the largest open-source robotics ecosystem, created in 2007 by the robotics incubator Willow Garage \cite{ros2009icra}.
It consists of a middleware and extensive tooling commonly required to build a variety of robotic systems.
However, it was built for researchers by researchers and lacked some features necessary to build robust commercial robots.

\textbf{ROS~2} was entirely redesigned to address the challenges facing the commercialization of robotics technology, while continuing its legacy of a broad ecosystem of distributed, community driven projects.
It tackles long-standing problems including security, embedded and real-time support, and operating in challenging networking environments~\cite{science}. 

ROS~2 also introduces approaches enabling developers to create performant, deterministic, and safe systems.
It encourages users to take control over a program's lifecycle, execution model, and resources. 
Two of the provided mechanisms are new types of \textit{Nodes} and \textit{Executors}.
\textit{Lifecycle} and \textit{Composition} nodes allow programs to have structural control over when and where a program executes, respectively \cite{science}.

This paper will focus on the latter, the composition node, which enables a designer to manually or dynamically instantiate nodes into processes.
They are also referred to more succinctly as \textit{Components}.
Composition was created to combat performance issues of one program per process while being unobtrusive to end-users \cite{science}.
Since many independent components may be assembled into a single process, ROS~2 maintains a distributed development model.
Composition \textit{dramatically} improves performance of the system from compute and memory, to latency and throughput.
The impact is significant when developing resource-constrained systems, practical systems with dozens of nodes, or particularly large messages common to sensor processing pipelines.
Indeed, processing pipelines with message sizes larger than 1 MB, such as images and point clouds, are virtually impossible without node composition.
Further, according to iRobot and discussed in Sec. \ref{irobot}, products like the iRobot Create{\sffamily\textregistered} 3 would not have been possible without ROS~2 composition reducing the computational burdens on its embedded processor. 
Composition also allows for intra-process communication (IPC), further improving performance with zero-copy. 

ROS~2 facilitates control over the execution model of the communication system.
These executors are specified in a program and define how information is processed within a node.
Thus, it is impossible to decouple the executor from the performance of a node - particularly when components are contained in the same process potentially sharing an executor.

In this paper, we briefly describe the concepts and design of process management in ROS~2.
We also introduce for the first time the executors and their complementary component containers needed to perform dynamic composition.

The paper's contribution is extensive performance benchmarking and analysis of composition to clearly demonstrate its power and importance within the roboticists' portfolio of tools.
These experiments are performed across multiple methods of composition, executor models, and communication.
Additionally, an illustrative experiment is performed on a mobile robot system using the popular ROS~2 Nav2 navigation system to reveal the significant system-wide improvements composition can make on an autonomous robot~\cite{nav2}.

\section{Related Work}
\label{sec:relatedwork}

There has been much research interest in the performance of the ROS~2 middleware framework due to its widespread use.
There have comparisons between ROS~1 and ROS~2's performance, as well as benchmarking the performance when varying the underlying DDS implementations \cite{maruyama2016exploring}.
Several efforts have created performance measuring tools for various facets of a ROS~2 system.
The {\tt ros2\_tracing} framework was introduced based on Linux Trace Toolkit Next Generation (LTTng) that allows for detailed analysis of instrumented events in a running ROS~2 system \cite{ros2tracing}.
Tracing analysis is extended by adding the ability to do message flow analysis by associating causally-linked tracing events in the ROS~2 computation graph \cite{bedard2023message}. 
Autoware\_Perf builds upon the {\tt ros2\_tracing} framework to add several measurements, including end-to-end latency, intra-node latency, and inter-node latency \cite{autoware_perf}. In addition, the tracing framework is evaluated against Autoware.Auto as a reference system, containing an architecture of nodes which simulates a real-world self-driving vehicle software scenario.

The Chain-Aware ROS Evalution Tool (CARET) also builds upon the {\tt ros2\_tracing} framework. CARET is used to perform analysis of executor performance against the Autoware reference system \cite{ros2scheduling, CARET}, but does not compare varying node architectures to characterize composition or differing communication methods (IPC, shared memory).  

In contrast to direct measurement using tracepoints, latency and determinism can be considered via formal analysis of the scheduling and execution of chains of callbacks.  For real-time applications, determining latency bounds is critical, so analysis to prove this is an important contribution \cite{casini2019response}. This analysis however does not account for inter-process versus intra-process communication, which was addressed in follow-on analysis \cite{processingchains}.  These analyses focus on worst-case latency of chains of callbacks, excluding composition and do not consider other key metrics. 

In addition, there is also work in improving the performance of ROS~2 by replacing the default executor.  One approach introduced a callback group executor with the aim to make ROS~2 suitable for real-time environments \cite{ros2realtime}. This executor improved latency and determinism.  This work did not study the use of composition or communication methods. Another approach to improving the executor is by introducing a scheduler that is aware of the the chain of callbacks to be executed \cite{picas}.  This work establishes the bounds on average and worst-case latency, but does not vary the communication mechanisms available in ROS~2.

While there is literature describing some elements of ROS~2 performance, our survey compares and analyzes the composition pattern which is not explicitly addressed in previous works. None of the existing body of work evaluates detailed performance metrics of composition and its applied relevance to robotics systems researchers and designers.

\section{Review of the Robot Operating System 2}
\label{sec:ros2}

\begin{figure}[ht]
    \centering
    \includegraphics[width=0.42\textwidth]{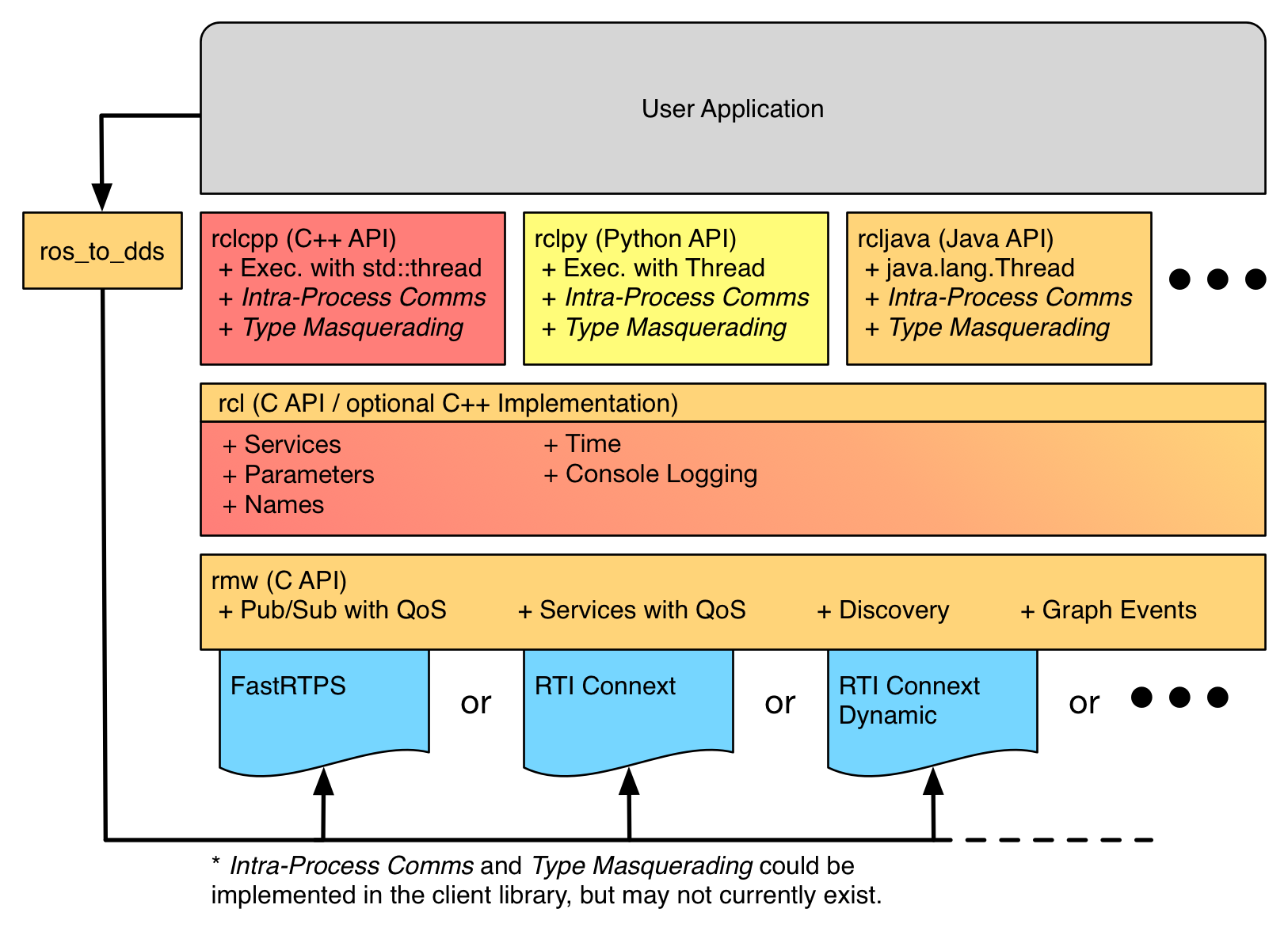}
    \caption{ROS~2's abstraction layers including DDS, {\tt rmw}, {\tt rcl}, and client libraries \cite{science}.}
    \label{fig:ros_cli}
\end{figure}

ROS~2 includes all of the familiar core features of ROS including publish/subscribe, services, actions, parameters, time, logging, and more.
Internally, ROS~2 is designed with modular abstraction layers to allow for use-case specific or updated integrations over time, shown in Fig. \ref{fig:ros_cli} \cite{science}.
The {\tt rmw} (ROS Middleware) is the abstraction for the middleware vendor to allow for multiple communication systems to be integrated.
The principal adoption of DDS for {\tt rmw} implementations has enabled ROS~2 to provide necessary features for modern robotics development \cite{science, dds}.
Though, non-DDS {\tt rmw}s exist with community supported implementations, such as {\tt rmw\_iceoryx} enabling low latency zero-copy communication \cite{ros2iceoryx}.
% DDS provides ROS~2 with improved performance in non-ideal networking scenarios such as those found in remote or highly congested environments using DDS's Quality of Service (QoS) dials \cite{dds}. 
Within ROS~2, there are two key concepts that largely define the behavior of a system above the {\tt rmw} - the Node and the Executor. 

\subsection{Nodes}
\label{sec:nodes}
A distributed system like ROS~2 can be conceptualized as a graph of end-points, each representing a process or subsystem.
The graph is built from these \textit{nodes} which act as units of processing.
% A distributed system like ROS~2 can be conceptualized as a graph of processing units called nodes, each representing a process or subsystem.
% Each node is an group of endpoints intended to serve a single, modular purpose in the context of the larger system.
These can be a hardware driver, data processing, localization, and more.
Each node connects to other nodes to communicate data via ROS interfaces; namely topics, services, or actions.
% Each node connects to other nodes to send data via one of the communication channels; namely topics, services, or actions.

There exist three major types of nodes. 
The most basic is the Node (without qualifiers) which is the base building block for all others. 
The next is the Lifecycle Node, which implements a state machine for managing the program deterministically for reliable execution. 
A node's state may be made dependent on other nodes to create a cascading bringup whereas each node is only activated once its constituent dependencies are also activated. 

Finally, the Composition Node allows for a program to be placed within processes either at compile or run-time.
This contrasts with other nodes, which are created in their own processes.
Composition reduces networking-related delays by using process-local optimizations and reduces overall resource consumption. 
This is explored in detail in Sec. \ref{sec:composition}.
Note that these nodes are not mutually exclusive; a node can be a lifecycle-component to leverage both deterministic execution and networking optimization. 

\subsection{Executors}
\label{sec:executors}

The \textit{Executor} is the mechanism by which the order and timing of available tasks are coordinated.
The executor allows the communication system's execution model to be decoupled from the structure of the system graph.
As network and timer events occur, they are marked in a structure called a \textit{waitset}.
The waitset is an array of entities, each indicating the status of the corresponding task.
% The waitset is a collection of entities, each of which is capable of generating asynchronous events.
% The waitset may be waited on, blocking until one of the entities produces an event.
Multiple nodes or groups of callbacks can be serviced by an executor.
When evaluated, an executor checks the waitset and services the next appropriate timer, topic, service, or action according to scheduling semantics, invoking relevant user-defined callbacks.

ROS~2's C++ client library currently provides three default executors.
The first is a Single-threaded executor, where all incoming events are serviced in order within a single thread.
A disadvantage of this executor is that any user-implemented blocking callback will also block other pending tasks.

As an alternative, the Multi-threaded executor uses a pool of threads to execute many user callbacks in parallel.
This can, however, cause callbacks to be starved if others use the shared resources greedily. 
To refine the parallelism, interfaces can be added to a \textit{callback group}.
Each callback group can be mutually exclusive (members should not be executed in parallel) or reentrant (members may be executed in parallel).

An executor's overhead can be considerable when reconstructing the waitset every iteration.
Thus, a third \textit{Static} single-threaded executor is available.
The static single-threaded executor does not reconstruct the list of event sources each iteration, but modifies an existing list as entities are added or removed.

In addition to the default set of executors, additional executors have been proposed: The \textit{Priority-Driven Chain-Aware Scheduler} \cite{picas} improves end-to-end system latency, the \textit{rclc Executor} \cite{rclcexec} provides deterministic scheduling for resource limited platforms, and the \textit{events-based executor} \cite{eventexec} eliminates overhead found in default implementations' waitsets.
% These alternative executors may also be applied to meet various application needs for determinism and latency, they may be used in conjunction with composition.

\subsection{Zero-Copy Communication}
\label{sec:IPC}

ROS~2 allows for improved performance through zero-copy communication using IPC and loaned messages optimizations.
Both avoid the copying of data to and from the middleware layer, serialization and de-serialization, and interactions with the ROS interfaces. 

IPC is used to maximize message passing performance of components in the same process, achieving similar performance to functional programming while retaining pub-sub semantics and modular architectures.
When a subscription and publisher of a topic reside in the same process, the underlying middleware can be entirely short-circuited; messages are passed via smart pointers pushed into shared ring buffers \cite{ros2consumerrobotics}.
IPC can be enabled through a node's options, but it presently has some implementation limitations: it is only compatible with a subset of Quality of Service (QoS) settings\cite{dds} and enabling it with subscriptions from other processes reduces its benefits.

The loaned messages API borrows memory from the middleware to perform optimizations like using shared memory buffers to enable multi-process zero-copy communication.
Using loaned messages requires the use of different publisher APIs and a relatively complex, vendor-specific configuration of the DDS layer and underlying Operating System to allocate the shared memory segment.
Messages published using loaned message must also have a fixed size that is known at compile-time.
If subscriptions from other machines exist, it is not recommended to enable loaned messages due to double delivery of messages.
Tests using loaned messages exhibit stability issues such as frequent crashes and dead-locks, discussed in further detail in Sec. \ref{sec:minimalexperiments}.
In contrast to IPC, these problems are due to the currently available {\tt rmw} implementations.
For these reasons, loaned messages support is considered experimental in ROS~2 as of the time of writing.
{\tt rmw\_iceoryx} is an example {\tt rmw} implementing shared-memory IPC with message loaning \cite{ros2iceoryx}.

\section{Overview of ROS 2 Composition}
\label{sec:composition}
%%  as it exists; it seeks to inform, not defend.
This section is a description of key details regarding composition as it exists important for understanding the contributed benchmarking and performance analysis in Sec. \ref{sec:minimalexperiments}. 
Composing multiple programs into the same process saves computing resources, reduces latency, and allows for a highly distributed development workflow to be employed.
However, this may not be optimal in all situations.
Thus, composition allows system designers to decide which process a program should reside in at either run-time or compilation-time without software changes.

For systems under development, components are often placed in separate processes so that failures do not disturb the larger system and analysis tools can be run in a well-defined scope.
As programs become more mature, they may be grouped into processes to reduce latency or share resources.

A key principle in the composition design is to be unobtrusive \cite{science}.
Unlike \textit{nodelets} in ROS~1, components do not require any specific method implementations and it is trivial to convert a typical node into a component. 
With a node class derived from {\tt Node}, the only requirement is that a constructor must take in a single argument, {\tt NodeOptions}, containing communication and node parameterizations.

There are two mechanisms for aggregating components: manual (compile-time) or dynamic (run-time) composition.
Note that either approach is available via the same API without code duplication.

\subsection{Manual Composition}
\label{sec:manual}
Manual composition is performed at compile time, with a user developed executable that explicitly instantiates components.
Since decisions are made at compilation-time, this method provides the least flexibility, but the maximum opportunity to control resource allocation between the components.
With manually-defined components in a process, the developer has the ability to explicitly control thread priorities and distribution.
Any executor(s) can be used in manual composition and is specified in the program's entry point.

In general, manual composition with a single executable is most optimal for encapsulating an entire system once all exploratory development is complete and the overall system architecture is static and stable.
Typically, this is leveraged in highly resource constrained environments, whereas minor overhead for flexibility is undesirable.

\subsection{Dynamic Composition}
\label{sec:dynamic}

Dynamic components are built as shared libraries and registered with a global index for bookkeeping.
They are then available to be loaded into a component container, which instantiates the nodes. 
This can be done within a ROS~2 launch file or from a service call.

In contrast to manual composition, dynamic composition retains maximum run-time flexibility, allowing designers to distribute components across various processes at or even after launch.
Distributing components allows for the grouping of related tasks in the same process, while retaining the ability to isolate components at will.
All components may be loaded into a single process, like in manual composition, with little overhead. 

\subsection{Component Containers}
\label{sec:containers}
Component containers provide a common executable for dynamically composing components, typically surrounding a particular executor.
The containers provide services to add, introspect, or remove components via the component manager.
A component container will search the component index and load the shared library module that contains a node of interest.
If the default containers are insufficient, new component containers may be written without modifying ROS~2 itself.

There are 3 default component containers available:
\begin{itemize}
    \item A single-threaded executor container, where components share the executor
    \item A shared multi-threaded executor container, where components share the executor
    \item An isolated singled-threaded executor container, where each component receives its own executor
\end{itemize}

\subsection{Best Practices}
\label{sec:use}

Best practices can be summarized as the following:
\begin{itemize}
    \item All nodes should be written as components
    \item Prefer dynamic composition over manual composition when not in extreme resource constrained environments
    \item With many independent components, prefer single-threaded executor(s) over one multi-threaded executor
    \item Prefer one multi-threaded executor when component(s) are specifically designed to not interfere with each other
\end{itemize}

\section{Benchmarking and Performance}
\label{sec:minimalexperiments}

The effectiveness of composition is evaluated in a variety of scenarios.
The following metrics were collected throughout the experiments:
\begin{itemize}
    \item Proportional Set Size (PSS): the physical RAM usage as the sum of its private memory and the fraction of its shared memory utilized divided by the number of processes it is shared with.
    This is the most accurate model as ROS~2 contains many shared libraries \cite{mackallpss}
    \item CPU percentage: the ratio between wall clock time and the time spent by the CPU to execute a process
    \item Latency: the time difference between message publication and a subscription callback invocation
    \item Goodput: application-level throughput, the amount of published data reaching the callback per second \cite{comernetworks}
\end{itemize}

Tests were run on Ubuntu 20.04.
The Linux {\tt proc} pseudo-filesystem was used to accurately record operating-system level metrics.
The communication-related metrics were computed by adding complimentary trackers in the source-code of the profiling applications.
The Humble Hawksbill distribution's default configuration were used throughout the experiments unless explicitly noted.
The CycloneDDS {\tt rmw} was selected for these experiments.
% All of the major {\tt rmw}s rely on the DDS protocol, e.g. CycloneDDS, Fast-DDS, and ConnextDDS.
% The {\tt rmw} options exhibit substantially similar trends, thus each are not reported due to the lack of additional insight~\cite{kronauer2021latency}.

All evaluations were performed on a Raspberry Pi 4 Model B, featuring an ARMv8-A 64 bit processor equipped with Quad-Core Cortex-A72 1.5 GHz CPU.
This embedded board was chosen because it is representative of the hardware commonly found in consumer robotics products.
Using a dedicated embedded test platform also reduced noise and jitter originating from background processes that could affect results.
Each experiment was conducted ten times and the measurements were averaged across consecutive runs.

We selected an existing framework\footnote{https://github.com/irobot-ros/ros2-performance} for reproducible evaluation of arbitrary ROS~2 systems, which we extended to support composition.

\subsection{Memory Footprint}
\label{sec:minimalmemory}

The first experiment evaluates the memory requirements of ROS~2 applications, comparing the two composition approaches with a standard multi-process system.
Each was tested with a varying number of empty nodes to evaluate baseline memory requirements.
The single-threaded executor was utilized for each test in this analysis.
Note that the choice of executor does not impact memory utilization.

\begin{figure}
\centering
\begin{tikzpicture}[scale=0.60]
\begin{axis}[
    ybar,
    bar width=6.5pt,
    xtick=data,
    symbolic x coords={1, 2, 3, 5, 10, 15, 20},
    xlabel={Nodes Number},
	ytick={10, 20, 30, 40, 50, 60, 70},
	ylabel={PSS [MB]},
    ymin=0,
    scaled ticks=false,
	legend pos=north west, % where is the legend
	legend image code/.code={\draw [#1] (0cm,-0.1cm) rectangle (0.2cm,0.25cm);}, % style of legend logo (the rectangle thingy)
	ymajorgrids=true, % show horizontal lines for Y ticks
	grid style=dashed, % style of Y ticks horizontal lines
    xtick style={/pgfplots/major tick length=0pt,},
]
% Multi-Process
\addplot[draw=RedLine, fill=RedFill] coordinates {
	(1, 9.53)
	(2, 11.62)
    (3, 13.69)
    (5, 18.31)
    (10, 31.66)
    (15, 50.27)
    (20, 72.13)
};
% Dynamic Composition
\addplot[draw=BlueLine, fill=BlueFill] coordinates {
	(1, 10.15)
	(2, 10.37)
    (3, 10.41)
    (5, 10.75)
    (10, 11.63)
    (15, 12.37)
    (20, 13.17)
};
% Manual composition
\addplot[draw=BrownLine, fill=BrownFill] coordinates {
	(1, 9.55)
	(2, 9.67)
    (3, 9.88)
    (5, 10.09)
    (10, 10.74)
    (15, 11.537)
    (20, 12.38)
};
\legend{Multi-Process, Dynamic Composition, Manual Composition}
\end{axis}
\end{tikzpicture}
\caption{RAM use for varying numbers of nodes.}
\label{fig:plotramvsnodes}
\end{figure}
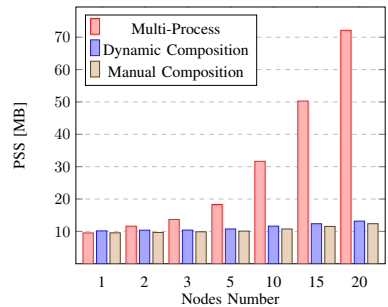

The results of this experiment are shown in \figref{fig:plotramvsnodes} and indicate a significant improvement utilizing any of the composition approaches.
The standard deviation between experimental runs was less than 1\%.

Each ROS~2 process creates its own DDS network participant to communicate, which represents a considerable memory increase.
The increase in memory in the multi-process system is non-linear: 2.1 MB increase on the second process that grows to 4.3 MB by the twentieth.
This is due to the creation of a fully connected and distributed network graph needing to allocate resources for every discovered end-point, growing $O(N^2)$.
This happens during network discovery and it is irrespective of whether the nodes will ever communicate.
The consequence is that a 20 node system with composition requires approximately the same memory as a 2 node multi-process system.
It is common for robot systems to contain 20 nodes or more; each hardware driver, robotic algorithm, or subsystem are typically modeled as a node. 
The autonomous navigation system evaluated in Sec. \ref{sec:nav2} alone contains 15 nodes, not counting the drivers, autonomy application, etc needed for a complete product.
Succinctly, practically complex robotic systems must use composition to stem the quadratic growth of DDS overhead.

Note that dynamically composed systems are slightly more expensive due to the additional component manager node.
They also require registration with an index and linking with the component library, both of which may be omitted for manual composition.
However, experiments showed a negligible increase in RAM (less than 100 KB) for twenty nodes.
Thus, it is recommended to register components to take advantage of flexible dynamic composition for prototyping and testing, even when somewhat resource constrained.

\subsection{CPU \& Latency Analysis for Different Message Sizes}

The next experiment evaluates the communication performance of composition.
This is not only affected by how the components are organized, but also by design decisions such as the selection of executors and communication settings.

The first test measures the steady-state CPU usage and mean latency across different systems comprising two nodes: one with a publisher and another with a subscription.
This experiment was run using both composition types and multi-process systems, each with a single-threaded executor.
The comparison also includes a dynamically composed system with IPC and a multi-process system using loaned messages.

Each test is repeated using different message sizes.
The publication frequency is set to 50 Hz to allow for appreciable differences without nearing limitations of the embedded platform for larger messages.
Subscription callbacks do not perform any computation, thus all of the measurements can be attributed to communication.

%TODO move to page 1 to be immediate value proposing
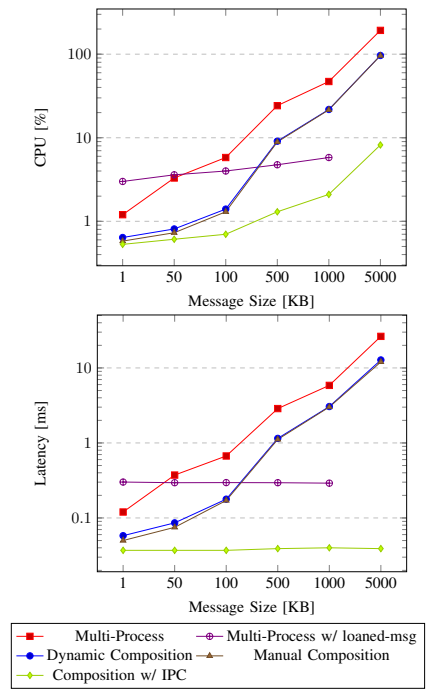
\begin{figure}
\centering
\begin{tikzpicture}[scale=0.60]
\begin{semilogyaxis}[
    xtick=data,
    symbolic x coords={1, 50, 100, 500, 1000, 5000},
    xlabel={Message Size [KB]},
    ylabel={CPU [\%]},
    ymajorgrids=true,
    grid style=dashed,
    log ticks with fixed point,
]
% Multi-Process
\addplot[red,every mark/.append style={solid,fill=red!80!black},mark=square*]
    coordinates {
    (1, 1.2)
    (50, 3.3)
    (100, 5.8)
    (500, 24.2)
    (1000, 47.1)
    (5000, 192.7)
};
% Multi-Process SHM
\addplot[violet,every mark/.append style={solid,fill=violet!80!black},mark=oplus]
    coordinates {
    (1, 3)
    (50, 3.6)
    (100, 4)
    (500, 4.75)
    (1000, 5.8)
    %(5000, ?)
};
% Dynamic Composition
\addplot[blue,every mark/.append style={fill=blue!80!black},mark=*]
    coordinates {
    (1, 0.64)
    (50, 0.81)
    (100, 1.4)
    (500, 9.1)
    (1000, 21.8)
    (5000, 96.5)
};
% Manual Composition
\addplot[brown!60!black,every mark/.append style={solid,fill=brown!80!black},mark=otimes*black,mark=triangle*]
    coordinates {
    (1, 0.58)
    (50, 0.73)
    (100, 1.3)
    (500, 8.9)
    (1000, 21.6)
    (5000, 96.1)
};
% IPC
\addplot[lime!80!black,every mark/.append style={fill=lime},mark=diamond*]
    coordinates {
    (1, 0.53)
    (50, 0.61)
    (100, 0.7)
    (500, 1.3)
    (1000, 2.1)
    (5000, 8.2)
};
\end{semilogyaxis}
\end{tikzpicture}

\begin{tikzpicture}[scale=0.6]
\begin{semilogyaxis}[
    xtick=data,
    symbolic x coords={1, 50, 100, 500, 1000, 5000},
    xlabel={Message Size [KB]},
    ylabel={Latency [ms]},
    legend style={at={(0.39,-0.2)},anchor=north,legend columns=2},
    ymajorgrids=true,
    grid style=dashed,
    log ticks with fixed point,
]
% Multi-Process
\addplot[red,every mark/.append style={solid,fill=red!80!black},mark=square*]
    coordinates {
    (1, 0.120)
    (50, 0.373)
    (100, 0.670)
    (500, 2.871)
    (1000, 5.852)
    (5000, 26.350)
};
% Multi-Process SHM
\addplot[violet,every mark/.append style={solid,fill=violet!80!black},mark=oplus]
    coordinates {
    (1, 0.301)
    (50, 0.295)
    (100, 0.296)
    (500, 0.295)
    (1000, 0.291)
    %(5000, ?)
};
% Dynamic Composition
\addplot[blue,every mark/.append style={fill=blue!80!black},mark=*]
    coordinates {
    (1, 0.058)
    (50, 0.086)
    (100, 0.178)
    (500, 1.150)
    (1000, 3.057)
    (5000, 12.771)
};
% Manual Composition
\addplot[brown!60!black,every mark/.append style={solid,fill=brown!80!black},mark=otimes*black,mark=triangle*]
    coordinates {
    (1, 0.050)
    (50, 0.075)
    (100, 0.171)
    (500, 1.114)
    (1000, 3.013)
    (5000, 12.025)
};
% IPC
\addplot[lime!80!black,every mark/.append style={fill=lime},mark=diamond*]
    coordinates {
    (1, 0.037)
    (50, 0.037)
    (100, 0.037)
    (500, 0.039)
    (1000, 0.040)
    (5000, 0.039)
};

\legend{Multi-Process, Multi-Process w/ loaned-msg, Dynamic Composition, Manual Composition, Composition w/ IPC}
\end{semilogyaxis}
\end{tikzpicture}

\caption{CPU, latency for different message sizes.}
\label{fig:plotcpulat}
\end{figure}

The results are shown in \figref{fig:plotcpulat} on a log-scale, showing reduction in both CPU and latency by more than half for every message size with either type of composition when compared to the standard multi-process approach.
Without zero-copy, performance degrades proportional to the message size.

Multi-process communication without loaned messages has a latency exceeding 30\% of the publication period (a key performance indicator) for messages above 1 MB.
Enabling loaned messages shows a near constant performance regardless of the message size.
This displays the highest overhead for small messages, but later demonstrates a major improvement for the larger messages - although overall remaining less efficient than IPC.
The loaned messages optimization showed problems when using messages larger than 2 Mb, which caused the application to repeatedly crash.
For this reason, this is not available in Fig. \ref{fig:plotcpulat}.

Each of the composition methods without IPC have remarkably similar performance; nevertheless the manual approach is slightly better due to the absence of the component manager.
Although this node does not process work at steady-state, the single-threaded executor is required to check it, adding a minor amount of overhead.
Indeed, the performance difference between the composition methods is only perceptible for the smallest message sizes where the manager contributes a larger proportion of the work completed \footnote{Later experiments show the `worst-case' (dynamic) due to the very small performance differences, adding clarity to later figures. Manual composition experiments were performed and exhibited analogous behavior.}.
% More advanced ROS~2 executors have been recently presented and, among other features, allow zero overhead for entities that do not do work \cite{rosexecutorworkshop}.

%STEVE
IPC in dynamic composition displayed a constant latency of 40 $\mu$s throughout all of the experiments, without scaling by message size.
CPU is reduced by almost an order of magnitude for larger messages; though it scales by message size due to the time it takes to populate the messages themselves.
Composition and IPC allow ROS~2 systems to achieve negligible latency, similar to monolithic applications.
Without Composition with IPC, ROS~2 could not support extreme resource constrained systems.

% a latency orders of magnitude lower than periods typically exploited in robotics, thus making it almost negligible.
%%%%%%%%%%%%%

\subsection{Maximum Achievable Goodput}

\begin{figure}
\centering
\begin{tikzpicture}[scale=0.60]
\begin{axis}[
    xtick=data,
    symbolic x coords={50, 100, 500, 1000, 5000},
    xlabel={Message Size [KB]},
    ylabel={Goodput [MB/s]},
    legend style={at={(0.39,-0.2)},anchor=north,legend columns=2},
    ymajorgrids=true,
    grid style=dashed,
    %log ticks with fixed point,
]
% Multi-Process
\addplot[red,every mark/.append style={solid,fill=red!80!black},mark=square*]
    coordinates {
    %(1, 12.13)
    (50, 167.74)
    (100, 165.3)
    (500, 156.77)
    (1000, 136.71)
    (5000, 126.13)
};
% Multi-Process SHM
\addplot[violet,every mark/.append style={solid,fill=violet!80!black},mark=oplus]
    coordinates {
    %(1, ???)
    (50, 21.97)
    (100, 43.94)
    (500, 219.72)
    (1000, 439.45)
    (5000, 540.18)
};
% Composition
\addplot[blue,every mark/.append style={fill=blue!80!black},mark=*]
    coordinates {
    %(1, 18.37)
    (50, 623.87)
    (100, 922.71)
    (500, 658.35)
    (1000, 541.17)
    (5000, 633.91)
};
% Composition IPC
\addplot[lime!80!black,every mark/.append style={fill=lime},mark=diamond*]
    coordinates {
    %(1, 23.38)
    (50, 978.51)
    (100, 1667.97)
    (500, 2875.57)
    (1000, 2980.60)
    (5000, 3064.31)
};
\legend{Multi-Process, Multi-Process w/ loaned-msg, Composition, Comp. w/ IPC}
\end{axis}
\end{tikzpicture}
\caption{Maximum goodput between one publisher and one subscription for different message sizes.}
\label{fig:plotgoodput}
\end{figure}
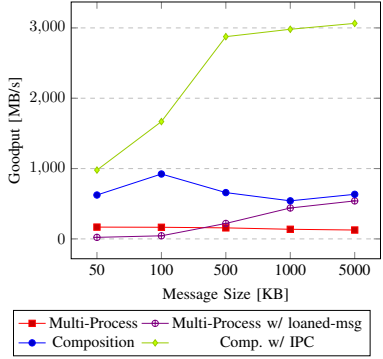

The next experiment computes the maximum amount of data that can be transferred between two entities.
The system is made up of a publisher and a subscription node, where we measure the goodput by publishing continuously without pause using a single-threaded executor.
This metric measures only the amount of useful data received by the subscription callback, excluding additional payloads used by the {\tt rmw} for the transmission.
The results for variable message sizes are shown in \figref{fig:plotgoodput}, where single-process solutions can reach a far higher maximum rate.

There is consistent overhead independent of data size from the executor and data handling deep within ROS~2.
For this reason, smaller messages have a relatively low goodput as these fixed costs reduce the maximum rate at which a subscription can receive data.

Multi-process without loaned messages reaches its peak for messages slightly smaller than 65 KB, which corresponds to the maximum UDP datagram length \cite{fall2011tcp}. 
When the payload exceeds that, the DDS middleware fragments the message into multiple packets for transport, with a negative impact on performance.
Although the multi-process approach transmits the least amount of data, it uses the most CPU, slightly above 200\%.
By comparison, a single-process approach used 82\%.

Using loaned messages with multi-process communication resulted in deadlocks and other stability issues when the publication frequency reached 500 Hz.
This required artificially limiting the maximum  rate to 450 Hz to obtain valid data and is why loaned messages underperformed when publishing small payloads.
The 1000 KB and 5000~KB tests did not reach the 450 Hz cap, with maximum frequencies 430~Hz and 110 Hz respectively. This allowed appreciable performance improvements to be seen while using loaned messages in the multi-process scenario at these message sizes - though still far slower than IPC.

Enabling IPC yields again the best results.
The goodput increases with message size, peaking due to the publisher's message allocation speed limit.

Our analysis shows that composition with IPC is a requirement for applications that must produce data at very high frequencies, such as above 1 KHz.
Many SLAM and Visual Odometry systems require images at 30+ Hz: using a VGA three channel image, multi-process cannot communicate an uncompressed feed at 20 Hz, while composition and IPC can stream at 85 Hz and 415 Hz, respectively.
Multi-process solutions with loaned messages can achieve the required frame-rate, but with a higher CPU usage, added complexity and, as of today, with the risk of stability issues.

%%%%%%%%%%%%%%

\subsection{Performance Impact of Component Containers}

\begin{figure}
\centering
\begin{tikzpicture}[scale=0.5]
\begin{axis}[
    xtick=data,
    symbolic x coords={1, 5, 10},
    xlabel={Subscriptions Number},
    ylabel={CPU [\%]},
    ymajorgrids=true,
    grid style=dashed,
    legend style={at={(0.3,0.9)},anchor=north},
]
% 1 EXECUTOR PER PROCESS NO WORK
%\addplot[blue,densely dashed,every mark/.append style={fill=blue!80!black},mark=*]
%    coordinates {
%    (1, 8.6)
%    (5, 13.4)
%    (10, 20.3)
%};
% 1 EXECUTOR PER PROCESS 500 WORK
\addplot[blue,every mark/.append style={fill=blue!80!black},mark=*]
    coordinates {
    (1, 10.9)
    (5, 25.8)
    (10, 44.7)
};
% ISOLATED (1 EXECUTOR PER NODE) NO WORK
%\addplot[red,densely dashed,every mark/.append style={solid,fill=red!80!black},mark=square*]
%    coordinates {
%    (1, 4.7)
%    (5, 33.6)
%    (10, 68.5)
%};
% ISOLATED (1 EXECUTOR PER NODE) 500 WORK
\addplot[red,every mark/.append style={solid,fill=red!80!black},mark=square*]
    coordinates {
    (1, 6.8)
    (5, 46)
    (10, 91.4)
};
% 1 MT EXECUTOR PER PROCESS NO WORK
%\addplot[brown!60!black,densely dashed,every mark/.append style={solid,fill=brown!80!black},mark=triangle*]
%    coordinates {
%    (1, 8.4)
%    (5, 37.4)
%    (10, 77.6)
%};
% 1 MT EXECUTOR PER PROCESS 500 WORK
\addplot[brown!60!black,every mark/.append style={solid,fill=brown!80!black},mark=otimes*black,mark=triangle*]
    coordinates {
    (1, 10.6)
    (5, 45.4)
    (10, 96.7)
};
\legend{SingleThreaded, Isolated, MultiThreaded}
\end{axis}
\end{tikzpicture}
\begin{tikzpicture}[scale=0.5]
\begin{axis}[
    xtick=data,
    symbolic x coords={1, 5, 10},
    xlabel={Subscriptions Number},
    ylabel={Latency [ms]},
    ymajorgrids=true,
    grid style=dashed,
]
% 1 EXECUTOR PER PROCESS NO WORK
%\addplot[blue,densely dashed,every mark/.append style={fill=blue!80!black},mark=*]
%    coordinates {
%    (1, 1092)
%    (5, 1570)
%    (10, 2243)
%};
% 1 EXECUTOR PER PROCESS 500 WORK
\addplot[blue,every mark/.append style={fill=blue!80!black},mark=*]
    coordinates {
    (1, 1.065)
    (5, 2.612)
    (10, 4.557)
};
% ISOLATED (1 EXECUTOR PER NODE) NO WORK
%\addplot[red,densely dashed,every mark/.append style={solid,fill=red!80!black},mark=square*]
%    coordinates {
%    (1, 647)
%    (5, 1751)
%    (10, 2523)
%};
% ISOLATED (1 EXECUTOR PER NODE) 500 WORK
\addplot[red,every mark/.append style={solid,fill=red!80!black},mark=square*]
    coordinates {
    (1, 0.630)
    (5, 1.870)
    (10, 3.137)
};
% 1 MT EXECUTOR PER PROCESS NO WORK
%\addplot[brown!60!black,densely dashed,every mark/.append style={solid,fill=brown!80!black},mark=triangle*]
%    coordinates {
%    (1, 1058)
%    (5, 1889)
%    (10, 3370)
%};
% 1 MT EXECUTOR PER PROCESS 500 WORK
\addplot[brown!60!black,every mark/.append style={solid,fill=brown!80!black},mark=otimes*black,mark=triangle*]
    coordinates {
    (1, 1.052)
    (5, 1.830)
    (10, 3.567)
};
%\legend{SingleThreaded, SingleThreaded w/ Work, Isolated, Isolated w/ Work, MultiThreaded, MultiThreaded w/ Work}
\end{axis}
\end{tikzpicture}

\caption{CPU, latency of single-process systems with varying executors. Subscriptions callbacks do 500~$\mu$s of work.}
\label{fig:plotexecutors}
\end{figure}
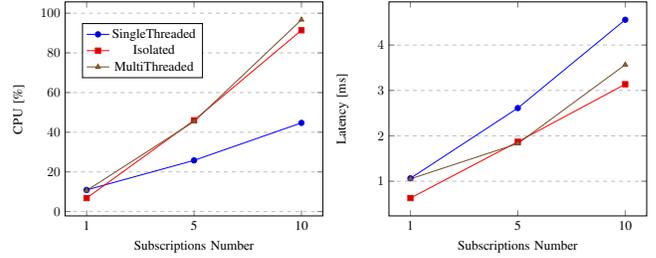

The component containers presented in \secref{sec:containers} play a fundamental role in the performance of a system via different executor patterns.
In this last experiment, we compared the container's performance using a simple dynamic composition application.
Each system comprises one publisher node, publishing messages of 500 KB at 50 Hz with a variable number of subscriptions.
Different from the previous tests, computations of 500 $\mu$s were added to each subscription callback to emulate data-processing (\figref{fig:plotexecutors}).

The single-threaded pattern requires the least CPU because the pipeline needs to interact with the {\tt rmw} only once per iteration, regardless of the number of subscriptions.
However, it also displays the highest latency since callbacks cannot execute in parallel.
Interestingly, when removing work from the callbacks, this approach actually becomes the fastest since the sequential execution of callbacks is outweighed by the reduced {\tt rmw} overhead.

The isolated and multi-threaded containers have similar performance, but with worthwhile differences. 
The multi-threaded container spawns the maximum number of concurrent threads supported by the architecture, which are not tied to any specific interface.
With only one subscription, parallelization is not possible, so the performance of the multi- and single-threaded containers are essentially identical.

The performance of the isolated and multi-threaded containers are similar as the number of subscriptions increases, with parallelization allowing to achieve lower latency than purely sequential execution.
The isolated container however yields better performance after the number of subscriptions exceeds the number of threads in the multi-threaded container (e.g. number of cores).
Additionally, the isolated container achieves the best performance when ROS~2's timers are used for periodic tasks.
Timers must be checked at every iteration of the executor, so by using dedicated executors for each node, the overhead of a timer applies only to the executor which contains it.

This experiment highlights that executors behave differently depending on the topology of the system and the constituent node's particular implementations.
In general, the isolated container provides a reasonable default when composing systems of non-trivial nodes.
Application developers are encouraged to explore the different modes and develop their own containers with more advanced configurations, such as specifying different types of executors for the different nodes or thread priorities and CPU affinities.

\section{Benchmarking on Realistic Robotic Systems}

Additional experiments were performed to showcase the practical impacts that ROS~2 composition has on autonomous robotics systems. 
The previous experiments clearly demonstrates the trends and behaviors of composition at a benchmarking level, but they do not illustrate the impact that could be expected on a full system when exploited.
Thus, this subsection is designed to briefly highlight the impacts with and without use of composition on full systems.

\subsection{ROS~2 Navigation}
\label{sec:nav2}

The Nav2 project is used as the autonomous robotics system for analysis \cite{nav2}.
It consists of a configurable and extensible behavior-tree navigator with several modular servers to accomplish tasks such as planning, control, recovery, path smoothing, route following, and more.
Each of these servers is modeled as an independent component-lifecycle node that can be placed dynamically throughout the computation network.
Nav2 is an ideal candidate for evaluation, with 15 nodes following the best practices in ROS~2 style, API, and design. 

\begin{figure}[ht]
    \centering
    \includegraphics[width=0.45\textwidth]{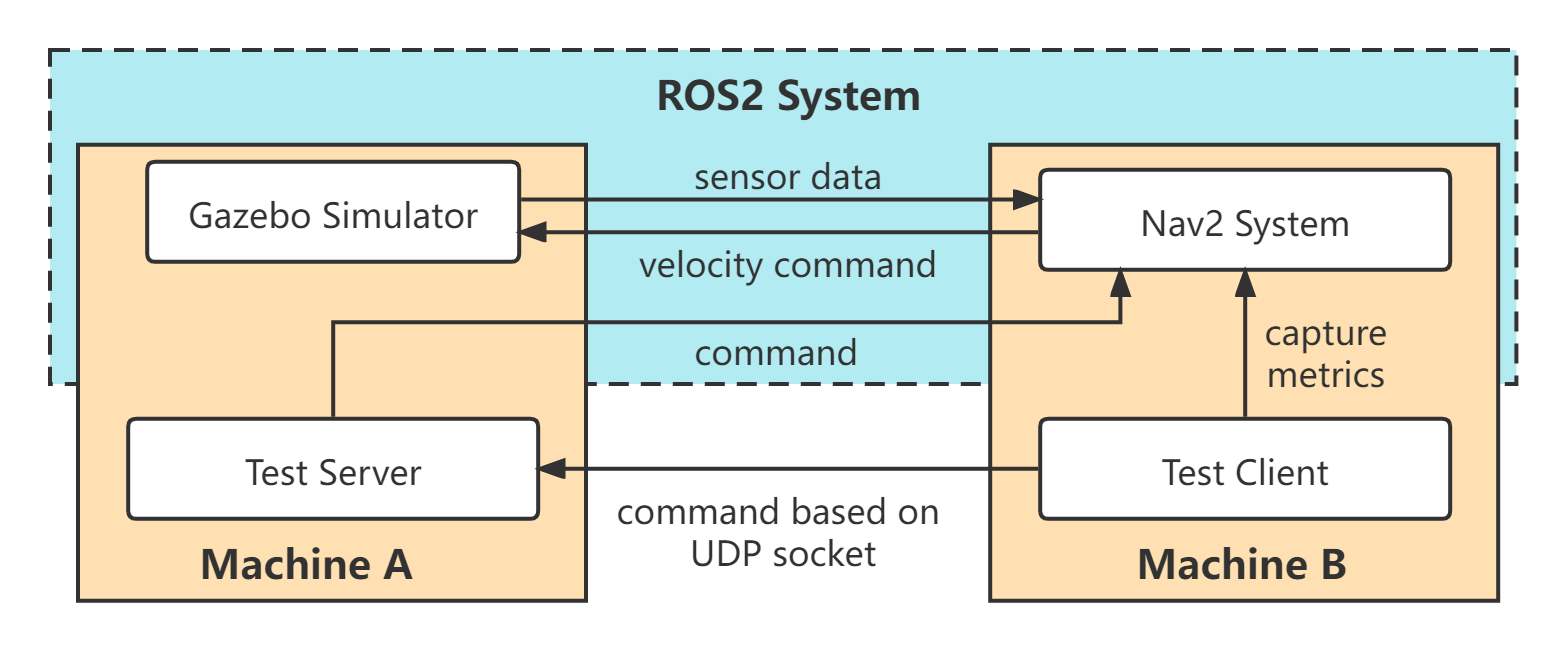}
    \caption{Testing framework for Nav2 system evaluation.}
    \label{fig:nav2testframework}
\end{figure}

This experiment captures the same computation and memory metrics on the same hardware platform as presented in Sec. \ref{sec:minimalexperiments}.
The navigation system is tested in the following configurations: multi-process, manual composition, and dynamic composition.
Both composition methods used the isolated single-threaded executor.

In these experiments, the navigation system was executed with a simulated differential-drive robot in Gazebo and measurements were taken during steady-state navigation between two arbitrary goal points on a map generated using SLAM Toolbox with sensor processing for dynamic obstacles \cite{Macenski2021, stvl, 1389727}.
It uses the NavFn global path planner alongside the DWB local trajectory planner processing inputs from an RGB-D sensor and a 2D planar laser scanner to represent a sandbox environment. 
The experimental setup and its configuration are identical to that which would be used in practical service robotics applications, which can be found in the {\tt nav2\_bringup} package. 

The testing scripts and simulator were run on a separate host computer to remove the impact of shared library components on PSS memory metrics, with setup in Fig. \ref{fig:nav2testframework}.
The results of this experiment are shown in Tab. \ref{nav2systemmetrics} representing the average of 10 trials.

\begin{table}[ht]
    \centering
    \caption{Nav2 resources used on ARM and x86 platforms.}
    \begin{tabular}{|l|c|c|c|}
    \hline
    \textbf{ARM} & \textbf{PSS [MB]} & \textbf{CPU [\%]}\\
    \hline
    Multi-Process                                                                                  & 116.63 ± 0.40 & 154.27 ± 3.91 \\ \hline
    Manual Composition                                                                            & 77.84 ± 0.47  & 110.50 ± 2.87 \\ \hline
    Dynamic Comp. (isolated) & 78.63 ± 0.17  & 109.62 ± 2.43 \\ \hline
    Dynamic Comp. (multi-threaded) & 75.52 ± 0.71  & 140.32 ± 7.05 \\
    
    \hline \hline
    \textbf{x86} & \textbf{PSS [MB]} & \textbf{CPU [\%]}\\
    \hline
    Multi-Process                                                                                  & 118.85 ± 0.36 & 48.60 ± 3.15 \\ \hline
    Manual Composition                                                                             & 67.13 ± 0.18  & 36.60 ± 4.22 \\ \hline
    \begin{tabular}[c]{@{}l@{}}Dynamic Comp. (isolated)\end{tabular} & 67.67 ± 0.14  & 36.09 ± 4.00 \\ \hline
    \begin{tabular}[c]{@{}l@{}}Dynamic Comp. (multi-threaded)\end{tabular}       & 66.71 ± 0.27  & 46.27 ± 2.80 \\
    \hline

    \end{tabular}
    \label{nav2systemmetrics}
\end{table}

Manual and dynamic composition on the embedded ARM processor both saved an astounding 28\% CPU and 33\% RAM over multi-process.
This is a resounding improvement in performance displayed by a full system containing advanced algorithms by composing them into a single process.
While both component containers consumed relatively similar memory, the multi-threaded container consumed far more CPU - on par with full-blown multi-process communication. 
This is due to the use of more nodes than CPU cores. 

A complementary experiment was run using an x86 platform (Intel i5-8300H) to highlight any differences due to compute architecture.
% Nav2, the testing scripts, and Gazebo were run on the same CPU, however they were confined in different docker containers to isolate performance metrics.
In that experiment, composition saved 25\% CPU and 44\% RAM over a multi-process system architecture - displaying equivalent trends on both major hardware platforms. 

These are significant improvements in performance due to only changes in the networking characteristics of an entire autonomous mobile robot navigation system. 
While continuing to plan, control, and perceive the environment, a robot has more resources to compute additional tasks such as its autonomy system or deploy more advanced perception techniques with the liberated memory and CPU time. 
Optionally, the freed memory resources could be used to store increasingly larger maps (40 MB can store an additional 100,000+ $m^2$) and freed compute time to plan longer global paths - enabling high quality navigation in larger scale spaces.
Over 10\% of an x86 core and 40\% of an ARM core are made available again for additional tasks which would not have otherwise been possible. 
Since navigation is only one system on a robot, it is essential to optimize its performance to leave room for other necessary subsystems (scheduler, external communication, error checking, payload application, data management, etc).

\subsection{iRobot Create{\sffamily\textregistered} 3}
\label{irobot}
The iRobot Create{\sffamily\textregistered} 3 is a robotic platform designed to give developers access to a robust mobile base through standard ROS~2 APIs.
It is equipped with a variety of sensors and actuators: encoders, optical flow and IMU for pose estimation, bumpers, cliffs and IR for obstacle detection, and wheels, LED lights and audio speakers controllable by the user.

The robot produces 70 KBps of raw sensor data and processed information, while also running a 50 Hz motion control loop and an obstacle detection pipeline.
Users can either publish individual actuation  commands or send action goals to execute autonomous behaviors such as pose regulation, wall-following, and docking.

The single-board computer that is present on the robot imposed severe resource constraints to the design of the application, due to a processor with limited CPU and less than 60 MB of RAM.
The robot runs a single-process, manually composed, ROS~2 application which takes advantage of IPC and an optimized executor \cite{eventexec}.
This ROS~2 system is made of approximately 10 nodes,  with more than 30 topics, 10 services and 10 action servers, in addition to the automatically created entities.
ROS~2 is used both for the internal implementation of algorithms and drivers, as well as for interfacing with the user.

The iRobot Create{\sffamily\textregistered} 3 application is normally controlled by an external navigation software, such as Nav2 described in Sec. \ref{sec:nav2}, and this results in the use of approximately 60\% CPU and 32 MB RAM.
More performance intensive operations such as remotely subscribing to all the published topics (e.g. for visualization purposes or to record a log), can increase the CPU usage to 80\%.
The robot also supports multiple {\tt rmw}, and with some of them the RAM usage grows up to 40 MB.

By taking into account the current resource usage and the results presented in the previous experiments, it is evident a multi-process solution wouldn't have been possible.
Indeed, in \figref{fig:plotramvsnodes} it is shown how a 10 nodes multi-process system required roughly 3 times the RAM of a single-process, composed, one - far exceeding the embedded compute platform's resources.
The run-time overhead of a multi-process configuration would have also caused consistent thread starvation and a CPU load average \cite{walker2006examining} above the number of available cores, making it impossible to achieve desired rates and smooth robot actions.
Without access to ROS~2 composition, the software architecture of this robot would have been vastly different, with a non-ROS monolithic application and a single ROS node acting as the bridge with the outside world.

When creating consumer products there is often a trade-off between cost and value.
The use of ROS does not directly add value to iRobot's customers; its main benefit is during development.
Thanks to ROS~2 composition, it is possible to deploy ROS-based applications on low-cost embedded platforms, as exemplified by the iRobot Create{\sffamily\textregistered} 3.

\section{Conclusion}
\label{sec:conclusion}

This paper presents how components and executors can highly impact ROS~2's performance with little overhead to a system designer.
Composition enables not previously possible applications of ROS~2 in extreme resource constrained systems and sensor processing pipelines while improving performance across complex robotic systems.
We showed experiments that evaluate these improvements and conclude that using any form of composition significantly improves memory, latency, and CPU usage. 
When IPC is enabled, it further decreases CPU usage by an order of magnitude and introduces low levels of stable latency.
These performance trends lead us to recommend composition as a best practice. 

% \section*{Acknowledgements}

% The authors thank the team at Open Robotics and Michel Hidalgo of Ekumen Labs for their outstanding contributions.

\bibliographystyle{IEEEtran}
\bibliography{citations}

\end{document}

%% file: config.tex
\usepackage{cite}
\usepackage{amsmath,amssymb,amsfonts}
\usepackage{algorithm}
\usepackage{algpseudocode}
\usepackage{graphicx}
\usepackage{textcomp}
\usepackage{hyperref}
\usepackage{xcolor}
\usepackage{subfig}
\usepackage{multirow}

\def\BibTeX{{\rm B\kern-.05em{\sc i\kern-.025em b}\kern-.08em
    T\kern-.1667em\lower.7ex\hbox{E}\kern-.125emX}}